# Paving the Way Towards Kinematic Assessment Using Monocular Video: A Preclinical Benchmark of State-of-the-Art Deep-Learning-Based 3D Human Pose Estimators Against Inertial Sensors in Daily Living Activities

*Mario Medrano-Paredes[1,2] (ORCID: 0009-0008-2638-2429), Carmen Fernández-González[1] (ORCID: 0009-0006-8018-7769), Francisco-Javier Díaz-Pernas[1] (ORCID: 0000-0002-5625-9607), Hichem Saoudi[1] (ORCID: 0009-0005-5307-7049), Javier González-Alonso[1] (ORCID: 0000-0003-3309-0578), Mario Martínez-Zarzuela[1] (ORCID: 0000-0002-6866-3316)*

**Abstract:**

Advances in machine learning and wearable sensors offer new opportunities for capturing and analyzing human movement outside specialized laboratories. Accurate assessment of human movement under real-world conditions is essential for telemedicine, sports science, and rehabilitation. This preclinical benchmark compares monocular video-based 3D human pose estimation models with inertial measurement units (IMUs), leveraging the VIDIMU dataset containing a total of 13 clinically relevant daily activities which were captured using both commodity video cameras and five IMUs. During this initial study only healthy subjects were recorded, so results cannot be generalized to pathological cohorts. Joint angles derived from state-of-the-art deep learning frameworks (MotionAGFormer, MotionBERT, MMPose 2D-to-3D pose lifting, and NVIDIA BodyTrack) were evaluated against joint angles computed from IMU data using OpenSim inverse kinematics following the Human3.6M dataset format with 17 keypoints. Among them, MotionAGFormer demonstrated superior performance, achieving the lowest overall RMSE (9.27° ± 4.80°) and MAE (7.86° ± 4.18°), as well as the highest Pearson correlation (0.86 ± 0.15) and the highest coefficient of determination $R^2$ (0.67 ± 0.28). The results reveal that both technologies are viable for out-of-the-lab kinematic assessment. However, they also highlight key trade-offs between video- and sensor-based approaches including costs, accessibility, and precision. This study clarifies where off-the-shelf video models already provide clinically promising kinematics in healthy adults and where they lag behind IMU-based estimates while establishing valuable guidelines for researchers and clinicians seeking to develop robust, cost-effective, and user-friendly solutions for telehealth and remote patient monitoring.

[1] Department of Signal Theory, Communications and Telematics Engineering. University of Valladolid, 47011 Valladolid, Spain.
[2] E-mail: mario.medrano@estudiantes.uva.es

# 1. Introduction

## 1.1. Background and motivation

Monocular Human Pose Estimation (HPE) is one of the most direct and promising applications in the field of computer vision. This discipline focuses on identifying, classifying, and tracking keypoints corresponding to human body joints using images and videos captured by digital acquisition systems. Together with inverse kinematics and biomechanical research, pose estimation has become a critical components in a wide range of applications, including sports performance analysis (Badiola-Bengoa & Mendez-Zorrilla, 2021), physical rehabilitation and healthcare (Avogaro et al., 2023). Over the past decade, the rise of telemedicine and digital health solutions has further underscored the importance of remote and cost-effective methods to evaluate movement quality in patients' everyday environments (Mizuochi et al., 2024). Telerehabilitation stands out as a way to remotely monitor patients' motor functions and physical activity beyond traditional clinical settings, thereby reducing hospital visits, enhancing patient adherence, and supporting long-term engagement in rehabilitative exercises (Seidman et al., 2017).

Current motion capture systems, such as multi-optoelectronic marker-based solutions (Vicon, OptiTrack, PhaseSpace, Qualisys) or multi-camera laboratory configurations, offer precise three-dimensional (3D) data but are often limited by high initial and operating costs, the need for fixed and controlled professional laboratory environments, and complex setup requirements. On the other hand, inertial measurement units (IMUs), which capture position, linear acceleration and angular velocity data, have emerged as a more portable solution: patients wear wireless sensors on different body segments, allowing for real-time precise monitoring outside the clinic. Nevertheless, these devices still pose challenges for both patients and practitioners, including the need for correct sensor placement and calibration (Höglund et al., 2021; Niswander et al., 2020), discomfort from wearing multiple units, which is even more pronounced in children and patients with specific physical conditions, and susceptibility to drift (Wittmann et al., 2019). Moreover, while sensor-based approaches have proven effective in controlled environments, they still introduce practical hurdles in day-to-day scenarios (Poitras et al., 2019).

Recent advances in computer vision and deep learning have promoted the development of video-based human pose estimation methods, which first appeared over a decade ago for example in (Ouyang et al., 2014; Toshev & Szegedy, 2014) offering an attractive, lower-cost alternative for telerehabilitation. By leveraging camera devices that are ubiquitous in smartphones, tablets, and laptops, these methods allow for an accessible yet efficient capture of body kinematics. In addition, this approach significantly reduces barriers to adoption since, instead of requiring specialized hardware and infrastructure, patients can perform exercises in natural home environments (Antón et al., 2013; Zhao & Krebs, 2024). Furthermore, powerful pose estimation algorithms—built upon convolutional neural networks (CNNs) (Fukushima, 1980), Transformers (Vaswani et al., 2023) or Graph Convolutional Networks (GCNs) (Kipf & Welling, 2017)—can automatically detect body landmarks and infer joint angles in real time, even in unstructured or partially constrained environments. This technology thus holds major potential for providing clinically relevant data and feedback at scale while minimizing costs and user burden. Building on these needs and constraints, we now review related work in monocular 3D human pose estimation that motivates our model selection and evaluation design.

## 1.2. Related work

Despite the significant progress made, there remains a notable gap in the literature concerning a direct, systematic comparison between deep learning-based human pose estimation and IMU-based methods for joint angle measurement in natural, out-of-the-laboratory conditions. Human pose datasets like Human3.6M (Ionescu et al., 2014), MPI-INF-3DHP (Mehta et al., 2017), Microsoft COCO (Lin et al., 2015) and 3DPW (Von Marcard et al., 2018) remain central to evaluating video-based HPE methods. These datasets illustrate diverse scenarios, including occlusions, multi-person interactions, and in-the-wild environments. Early works, such as DeepPose (Toshev & Szegedy, 2014), DeeperCut (Insafutdinov et al., 2016) or (Chakraborty & Namboodiri, 2017) pioneered video-based human pose estimation using DNNs and CNNs. While effective, these methods often underscored gaps in temporal consistency, depth ambiguity, accuracy errors and occlusions.

Comprehensive reviews, such as (Andriluka et al., 2018; Ben Gamra & Akhloufi, 2021; Y. Chen et al., 2020; Pang et al., 2022; J. Wang et al., 2021; C. E. Zheng et al., 2018) systematically analyze multiple deep learning methods and architectures for both 2D and 3D HPE, emphasizing challenges like occlusion and depth ambiguity in monocular video data. These reviews highlight progress in 3D pose estimation frameworks, particularly for applications in motion analysis and telerehabilitation. However, existing benchmarks often prioritize controlled laboratory settings, neglecting real-world challenges such as dynamic lighting and variable camera angles, inherent to rehabilitation in uncontrolled environments. In addition, only a few studies discuss the trade-offs in accuracy, inference time and ease of use between models, as well as the best specific applications depending on the type of architecture (K. Chen et al., 2018; Ienaga et al., 2022). Regarding the ground-truth systems, most of the existing reviews (Y. Lee et al., 2024; Vafadar et al., 2022) compare computer vision architectures to multi-optoelectronic marker-based configurations with several infrared cameras such as Vicon or Optitrack, which do not meet accessibility and budget requirements for home-based rehabilitation.

We distinguish two families of monocular 3D human pose estimation. End-to-end estimation (Mehraban et al., 2023) predicts 3D joint coordinates directly from RGB frames with a single network. These models can use image appearance to reason about occlusions and often learn temporal priors when trained with video, yet their generalization may degrade when target cameras and backgrounds differ from the training distribution. In contrast, 2D-to-3D pose lifting (T. Jiang et al., 2023) first detects 2D joint locations and then infers 3D pose from the two-dimensional sequence. Lifting depends on the quality of 2D detections, so localization errors can propagate to 3D, and temporal consistency is handled by modules such as temporal convolutions or attention over 2D pose sequences, which can be challenged by severe occlusions or prolonged non-repetitive motions.

Several new HPE methods have emerged in recent years in response to the rapid advances in the field of Machine Learning, Transformers being the most popular one. For example, ViTPose (Xu et al., 2022) introduced a streamlined baseline that leverages plain, non-hierarchical vision transformers for pose estimation without relying on additional CNN-based feature extractors. The model's design emphasizes scalability and training flexibility, achieving state-of-the-art results on MS COCO. Moreover, (C. Zheng et al., 2021) proposed a fully transformer-based model that directly estimates 3D human poses from 2D pose sequences, leading to state-of-the-art performance on benchmarks like Human3.6M and MPI-INF-3DHP.

However, other technologies have also been explored. (Z. Jiang et al., 2023) introduces ZeDO, a zero-shot diffusion-based optimization pipeline for 3D HPE that iteratively refines poses by minimizing 2D reprojection errors using a pre-trained diffusion model. Unlike conventional learning-based methods that struggle with domain shifts by averaging poses, ZeDO estimates each pose case-by-case without relying on paired 2D-3D training data. It achieves state-of-the-art performance with strong cross-domain generalization, as it combines an initial pose optimizer with iterative denoising steps to bridge the gap between optimization and learning-based approaches.

HPE has been leveraged in biomedical contexts for multiple tasks including gait analysis (Martini et al., 2022; Topham et al., 2023) and fall detection using OpenPose (Huang et al., 2018). In (Rosique et al., 2021), the authors present ExerCam, a low-cost telerehabilitation system that leverages a standard RGB webcam and the OpenPose 2D library to perform real-time human pose estimation. ExerCam supports both task-based evaluations: measuring range of motion through calibrated target exercises, and gamified modes designed to enhance patient engagement. Additionally, the system features a web-based management platform that enables therapists to remotely monitor patient progress and adjust therapy protocols accordingly. While some works have studied the feasibility of video-based telerehabilitation methods (Milosevic et al., 2020), there is no existing comparison between state-of-the-art HPE architectures oriented towards accessibility in clinical applications. Building on these advances and gaps, we now state the objectives and the contributions that guide our benchmark.

### 1.3. Objectives and contributions

There is a gap in the literature regarding a direct and clinically grounded comparison of accessible, monocular video-based 3D HPE against joint angles obtained from IMUs in real-world environments for daily living activities. Prior work often relies on optical motion capture that is not practical for home use, or lacks comparative analyses of model efficiency, daily movements, and validation in uncontrolled environments.

Our work addresses these gaps by conducting a preclinical benchmark on healthy adults evaluating state-of-the-art models—specifically MotionAGFormer (Mehraban et al., 2023), MotionBERT (Zhu et al., 2022), MMPose, and NVIDIA Maxine AR SDK BodyTrack—against IMU data. This study aims to provide quantitative insights into models' performance focusing on clinical metrics such as joint angle accuracy under an out-of-the-lab environment, laying the groundwork for future patient studies.

Therefore, our main objective is to provide a preclinical evaluation in healthy adults that quantifies how accurately current monocular video methods recover joint angles compared with angles derived from inertial sensors, and to extract activity-dependent insights that guide method selection and deployment in telerehabilitation and related applications. The specific aims of the study are first to establish a unified and reproducible pipeline that converts outputs from four monocular 3D HPE models into joint angle trajectories, using a harmonized Human3.6M joint set with 17 keypoints, a vector-based angle formulation, and a documented sequence of filtering and synchronization. Second, quantitatively compare those models against IMU-derived joint angles across 13 daily living activities recorded with healthy adults, a commodity camera and five IMUs. Third, to analyze activity-dependent trade-offs in accuracy, temporal agreement, and deployment considerations that matter for remote kinematic assessment. Fourth, to use evaluation metrics that are interpretable for kinematics—in particular RMSE, MAE, Pearson correlation, and $R^2$—computed on filtered and synchronized signals, and report both overall and activity-specific results.

Consequently, our most notable contributions are outlined below:

(i) A quantitative comparison of state-of-the-art deep neural network architectures for tracking human poses while performing daily living activities in real-world scenarios, validated against a more reliable 3D measure using IMU sensors.
(ii) An end-to-end, transparent and fully documented validation protocol that harmonizes model outputs to a common skeleton, shows the signal streams from raw inputs to metric computation, and improves methodological reproducibility.
(iii) Identification of strengths and weaknesses for each architecture with respect to expected clinical situations, including trade-offs between accuracy and latency, together with evaluation of deployment factors such as inference speed, ease of integration into existing workflows, and hardware requirements.
(iv) Actionable insights for researchers and clinicians that guide the design of future patient monitoring systems, including guidance on camera placement, model use depending on the joint under study, viewpoint, data processing, and overall feasibility for home-based use.
(v) Evidence of robustness in realistic partially uncontrolled conditions that helps bridge the gap between laboratory testing and future in-home clinical applications.

In addition to these aspects, we provide open research assets that include processed signals, figures, and tables in order to support independent verification and future extensions of the benchmark. With the aims and contributions defined, the next section details the dataset, the models, and the processing pipeline used to derive joint angle trajectories.

## 2. Methods and materials

This section describes the VIDIMU dataset, the video-based and IMU-based HPE pipelines, and the evaluation procedures used to compare the models.

### 2.1. VIDIMU dataset description

#### 2.1.1. Overview

The VIDIMU dataset (Martínez-Zarzuela et al., 2023) is a multimodal resource focused on daily-life activities that comprises video recordings of 54 healthy adults (36 males, 18 females; age 25.0 ± 5.4 years) performing 13 clinically relevant daily-life activities, with 16 subjects simultaneously recorded using video and Inertial Measurement Units (IMUs). Activities are categorized into lower- and upper-limb tasks. The first ones include walking forward and backward, line walking and sit to stand transitions (A01-A04), while the second ones range from simple movements such as bottle manipulation to complex bimanual actions like LEGO assembly and ball throwing (A09-A13).

#### 2.1.2. Video acquisition setup and protocol

Detailed information on the activities performed, their development and the number of repetitions can be found in the original VIDIMU paper. The protocol stresses that a patient's position and orientation relative to the camera must be chosen according to the task type and whether it targets upper or lower extremities, as these factors directly affect HPE accuracy and robustness to occlusions. Having established the data and acquisition protocol, we next introduce the 3D human pose estimation models evaluated in this study.

### 2.2. 3D human pose estimation deep learning video models

Four state-of-the-art 3D HPE models have been employed to infer poses from key joint points detected from video recordings (see Fig. 1). Their differences in architecture and training strategies provide valuable insights into the trade-offs between efficiency, accuracy, and adaptability in human motion analysis.

The benchmark includes four monocular 3D HPE systems chosen to represent the two dominant families in the literature and to cover a range of deployment profiles. MotionAGFormer and MotionBERT are recent methods that recover 3D poses with strong temporal modeling. MMPose is a practical three stage lifting pipeline that combines 2D detection, 2D pose estimation, and temporal 2D to 3D lifting within a single framework. NVIDIA BodyTrack is a close source end-t- end system that predicts 3D joints directly from RGB and is included for continuity with the original VIDIMU paper and for its relevance in applied single camera scenarios. This selection balances accuracy, openness, and real-world usability while keeping all outputs compatible with the Human3.6M joint set used downstream for angle computation. Detailed descriptions of each system is detailed below.

For open-source models we used the official pretrained weights and default inference configurations unless otherwise stated. Some implementation details and parameter counts were not consistently reported across projects. NVIDIA BodyTrack is a closed-source SDK, so internal architecture, training data, and parameter counts are not publicly disclosed. We therefore reported all verifiable settings for each model.

### 2.2.1. MotionAGFormer

MotionAGFormer (Mehraban et al., 2023) is a hybrid model that combines the advantages of transformer-based self-attention mechanisms with Graph Convolutional Networks (GCNs) (Kipf & Welling, 2017) to enhance the accuracy and efficiency of 3D HPE. Traditional transformer-based (Vaswani et al., 2023) models struggle with capturing fine-grained local dependencies between adjacent joints while excelling at modeling long-range relationships. Conversely, GCNs are well-suited for local feature extraction but are limited in their ability to represent global dependencies. To address this, MotionAGFormer introduces the Attention-GCNFormer (AGFormer) block (B. Jiang et al., 2023), a dual-stream architecture where one component processes spatial and temporal dependencies using Transformers, while the other employs a graph-based representation to capture local relationships between joints. The outputs of these two parallel processing streams are adaptively fused to balance the integration of both local and global feature representations.

The model takes as input a sequence of 2D human poses detected from monocular (single-camera) video and projects each joint into a higher-dimensional feature space. Spatial and temporal embeddings are added to preserve positional information before passing the data through multiple AGFormer blocks, each of which applies transformer-based self-attention and GCN-based processing separately. MotionAGFormer predicts entire sequences of 3D poses rather than just the central frame, which allows it to leverage temporal coherence while minimizing computational redundancy. The final 3D pose predictions are refined using a regression head, and the training objective incorporates both position and velocity losses to improve smoothness and accuracy. The model achieves state-of-the-art results on the Human3.6M (Ionescu et al., 2014) and MPI-INF-3DHP (Mehta et al., 2017) datasets while being more computationally efficient than its predecessors. Additionally, it is offered in multiple variants (XS, S, B, L) to provide flexibility in balancing precision and performance.

### 2.2.2. MotionBERT

MotionBERT (Zhu et al., 2022) takes a different approach to 3D human pose estimation by focusing on pretraining a robust motion representation that can be transferred across multiple human-centric tasks. Instead of solely relying on supervised learning for pose lifting, it employs a self-supervised pretraining strategy in which a motion encoder is trained to reconstruct 3D poses from partial, noisy 2D skeleton sequences. This process forces the model to learn meaningful motion priors that encode geometric, kinematic, and physical constraints of human movement. The core of MotionBERT is the Dual-Stream Spatio-Temporal Transformer (DSTFormer) (Zhang et al., 2024), which consists of two separate branches: one processes spatial relationships between joints within a single frame, while the other captures temporal dependencies across frames. Unlike previous methods that stack spatial and temporal operations sequentially, it fuses the outputs of these two streams adaptively, ensuring that both local and global motion features contribute to the final 3D pose estimation.

This system is pretrained on large-scale human motion datasets, where 2D skeletons are corrupted with occlusions and noise to simulate real-world conditions. The model learns to recover the missing information and infer the depth of joints, resulting in a highly generalizable motion representation. During fine-tuning, the pretrained motion encoder is adapted for specific downstream tasks, including 3D pose estimation, action recognition, and human mesh recovery. This pretraining strategy significantly improves the performance of MotionBERT compared to models trained from scratch. The model achieves state-of-the-art results on the Human3.6M dataset, demonstrating lower mean per joint position error (MPJPE) and superior temporal smoothness in pose predictions. Due to the clinical orientation of our work, and the focus on the analysis of the human body, the main metric used in our study is the computation of joint angles based on the estimated keypoints, instead of simply using the joint positions.

### 2.2.3. MMPose 3-stage 2D-to-3D lifting architecture

MMPose is a pose estimation open-source toolkit, based on PyTorch and part of the OpenMMLab project. Among the algorithms offered in this framework, there is a configuration representing a state-of-the-art three-stage architecture for video-based 3D human pose estimation. From now on, we will refer to this specific entire system as MMPose for simplicity of notation. Implementation details of each step are described below.

In the first stage, RTMDet (Lyu et al., 2022), an efficient convolutional neural network optimized for real-time detection, localizes human subjects within video frames using bounding box detection. This detector employs a modified CSPNet (C.-Y. Wang et al., 2019) backbone with cross-stage partial connections and a scaled feature pyramid network (FPN) (Lin et al., 2017) head, achieving high person detection accuracy at 640×640 resolution.

The second stage utilizes RTMPose (T. Jiang et al., 2023), a top-down 2D pose estimator that combines a HRNet-style (J. Wang et al., 2020) backbone with SimCC (Simulated Classification-based Coordinate Regression) head. This novel approach discretizes the continuous coordinate regression task into two independent classification sub-tasks for horizontal and vertical axes, enabling sub-pixel precision while maintaining computational efficiency.

The final stage employs VideoPoseLift, a temporal convolution network (TCN) (Lea et al., 2016) that lifts 2D keypoints to 3D joint positions through spatiotemporal analysis. The architecture processes 243-frame temporal windows using dilated causal convolutions with residual connections, capturing long-range dependencies while maintaining temporal causality. The model implements a fully convolutional design with 1D temporal convolutions across feature dimensions, enabling efficient processing of video sequences. This three-stage pipeline is trained in a supervised manner on large-scale datasets (Human3.6M (Ionescu et al., 2014), Microsoft COCO (Lin et al., 2015), with intermediate modules pretrained on domain-specific data to enhance robustness. The implementation leverages MMPose's optimized inference engine, which integrates model cascading and GPU-accelerated postprocessing for real-time performance.

### 2.2.4. NVIDIA Maxine AR SDK BodyTrack

Original VIDIMU (Martínez-Zarzuela et al., 2023) pipeline integrates BodyTrack, a proprietary 3D HPE from NVIDIA Maxine Augmented Reality (AR) software development kit (SDK). Initially, BodyTrack was selected for its unique capability to infer human joint 3D coordinates (x, y, z) from single-camera footage, a critical feature for real-world applications where multi-view setups are impractical. However, as a closed-source tool, NVIDIA Maxine's architecture, training data, and optimization strategies are undisclosed, limiting reproducibility and transparency. In addition, different open-source alternatives using the latest advances in computer vision have emerged in recent years.

While its performance in controlled experiments is validated by the dataset's benchmarks, the reliance on proprietary algorithms introduces constraints for academic scrutiny and adaptation. Furthermore, the lack of access to internal parameters or training methodologies raises questions about generalizability beyond NVIDIA's ecosystem, as well as its feasibility of use in clinical and healthcare settings. Despite these limitations, Maxine AR SDK integration into VIDIMU underscores its utility in bridging video-based pose estimation with clinical-grade motion analysis, albeit with inherent dependencies on NVIDIA's proprietary technology.

After describing the models, we explain how their raw outputs are converted into a unified format and into joint angle time series.

### 2.3. Video data processing

All video-based 3D HPE deep learning models in the benchmark follow a similar workflow for data extraction and preprocessing. The data was processed using a modular Python pipeline consisting of (i) HPE, video data processing, and file conversion, (ii) joint set harmonization, (iii) joint angle computation, (iv) linear interpolation, (v) filtering, and (vi) standardization. The end-to-end pipeline can be observed in Fig. 2. The specific process is described in detail below.

Each HPE system provides 3D joint coordinates for every frame, which are then converted to a unified comma-separated value (CSV) format to facilitate downstream analyses. The second step is joint set harmonization and raw pose file conversion. For each model. dedicated conversion scripts iterate through each frame's pose annotations, extract the 3D coordinates, and match each model's native joint naming convention to our target naming scheme (e.g., mapping *neck_base* to *neck*). The final CSV file likewise contains columns of {x,y,z} coordinates for each joint in each frame. Once converted, all CSV files share a consistent joint naming convention and column format. They are then read into a Pandas DataFrame.

Note that NVIDIA BodyTrack natively outputs 34 keypoints, while the rest of the models follow the Human3.6M dataset format, consisting of 17 keypoints. To achieve uniformity and ensure that all calculated angles belonged to the same joints regardless of the model used, this format was also used for this study. Thus, the 34 BodyTrack markers were mapped to the 17 existing ones via harmonization scripts, discarding the remaining 17. It was not necessary to apply any type of interpolation to the existing data.

Next, joint angle computation is performed in order to derive joint kinematics from 3D coordinates. First, a subset of the 3D markers is taken for each relevant joint (e.g., right elbow, right shoulder). Afterwards, two vectors are constructed, representing the bones around the joint of interest (e.g., upper arm and forearm for the elbow). Finally, the angle between those vectors is calculated using the dot product and the arccosine function, as shown in Eq. (1). The markers used to calculate each joint angle, as well as the joint angles calculated for each activity, appear in Table 1 and Table 2 respectively.

$$\theta(t) = \arccos\left(\frac{v_1(t) \cdot v_2(t)}{||v_1(t)||\,||v_2(t)||}\right) \times \frac{180}{\pi} \quad (1)$$

To convert from frame-based data to time-based data, the camera frame rate is used (in our case, 30 Hz as it is the frame rate for the cameras employed during VIDIMU dataset video acquisition). As a result, the CSV data for each frame can be transformed into an angle time series for upper- or lower-limb human body joints.

Because 3D HPE video deep learning models often contain missing or noisy values, two additional processing steps are performed. First, gaps or invalid entries are interpolated, thereby ensuring continuous angle signals. Second, a filtering stage is applied. A median filter with window size of 5 samples is applied to smooth the signals before plotting and remove peaks. A moving average filter is also applied to mitigate high-frequency and minor frame-to-frame jitter. These filtering steps are applied to the joint angle trajectories before any plotting and before computing the evaluation metrics. All reported metrics are calculated on these filtered signals. The window sizes for the median and moving average filters were selected experimentally. A range of plausible values was explored and iteratively adjusted while inspecting the resulting trajectories to balance noise attenuation and waveform fidelity. The final values were then fixed and used uniformly across all participants, activities, and models to ensure comparability.

As our scenario involves a comparison of the video-derived joint angles against inertial sensor measurements, which serve as the reference data, an additional standardization procedure is required. In brief, the angle signals are first normalized using mean removal and then shifted and truncated to align them with the reference signals by minimizing the root mean squared error (RMSE). This step is model-agnostic and applies the same synchronization logic to any CSV-based 3D pose estimation data.

The pipeline also incorporates a visualization module that generates plots of both the raw and processed signals in each step, as well as the comparison between synchronized video- and IMU-based angle joint signals. These visualizations (in SVG and PDF formats) allow for qualitative assessment of the filtering and synchronization steps. For example, the IMU joint angles (after filtering and resampling) and the corresponding video-based estimates are overlaid to verify alignment. All the files generated are available in Zenodo (Medrano-Paredes et al., 2025). This step served as a quality control measure to identify potential acquisition or processing errors.

Through this standardized pipeline, we ensure that all four HPE methods (BodyTrack, MotionBERT, MotionAGFormer, and MMPose) produce temporally and spatially comparable angle trajectories for our subsequent benchmarking and evaluation steps. In parallel, the IMU stream follows a homologous pipeline so that both modalities produce directly comparable angle trajectories.

### 2.4. IMU data processing

Although optical motion capture is commonly considered the in-lab gold standard, our target application is out-of-the-lab assessment where portability, low cost, and rapid setup are required. We therefore use IMU-derived joint angles from OpenSim inverse kinematics as the reference for comparison with video models. Prior comparative work reports good agreement for joint-angle estimation and highlights advantages of inertial sensors in cost, ease of use, and patient adherence in non-laboratory settings (Nijmeijer et al., 2023; Wilhelm et al., 2023). This justifies IMUs as a pragmatic reference for the present preclinical benchmark focused on remote kinematics, while we acknowledge that optical validation would strengthen future studies.

Raw inertial measurement unit (IMU) data were acquired using custom sensor hardware and then processed using OpenSim  Data were then stored in multiple formats (e.g., *.raw*, *.mot*, and *.sto*). For the present analysis, we focused on the joint angle estimates obtained via inverse kinematics (IK) in OpenSim. The data was processed using a modular pipeline similar to the one used for video-based angle signals, which comprises the following stages: (i) OpenSim IK and file conversion, (ii) downsampling to 30 Hz, (iii) linear interpolation, (iv) filtering, and (v) standardization. The complete workflow can be observed in the lower branch of the diagram shown in Fig. 2. The step-by-step process is detailed below.

After capturing the movement and poses using custom IMUs in parallel to video acquisition as described in VIDIMU, the quaternion data obtained (*.raw* files) were fed into OpenSim to perform the inverse kinematic (IK) computation of human body joint angles, obtaining a subset of *.mot* files as a result. In this case, it was not necessary to develop a proprietary implementation for the angle calculation, since OpenSim performs this analysis automatically and outputs the joint angles data in a *.mot* file.

First, missing data points were recovered using linear interpolation, ensuring that the video and IMU signals had continuous trajectories for subsequent synchronization. The raw IMU signals, originally sampled at 50 Hz, were downsampled to 30 Hz using a resampling procedure to match the frame rate of the video-based human pose estimation. In addition, the initially extracted joint angle signals were affected by occasional high-frequency noise and missing values. To address these issues and smooth the signal, a median filter (with window size N=5 samples) was applied. A moving average filter was subsequently employed to further smooth the signals.

To compare IMU-derived joint angles with video-based estimates, it was essential to temporally align the two modalities using a standardization procedure. After normalizing the signals using mean removal, an iterative synchronization procedure was implemented: for a fixed fitting window (180 samples corresponding to the first six seconds of video), the root mean squared error (RMSE) between the IMU and video signals was computed. The signals were then shifted relative to one another (up to a maximum offset of 15 samples) until the RMSE was minimized. The final synchronized signals were then trimmed to the maximum common length. Further implementation details can be found in the original VIDIMU work.

As in the case of the angle signals extracted from the video, the workflow also incorporates a section for visualization of both the raw and each step of the signal processing. The IMU angle signals and the angle signals for each 3D HPE model are overlaid so that alignment and correlation can be easily verified, as shown in Fig. **3**. Plots for the rest of the activities can be found in the Supplementary Materials section and in the companion Zenodo repository (Medrano-Paredes et al., 2025). The visualizations are also saved in both SVG and PDF file formats.

This processing pipeline underpins the subsequent benchmark analysis between video-based HPE deep learning models and IMU-based reference measurements. Its modular design ensures both flexibility and extensibility in processing multimodal motion data. Finally, we define the metrics and the statistical analysis used to quantify agreement between modalities.

**2.5.  Evaluation metrics and statistical analysis**

The performance of the different deep learning models for 3D HPE was evaluated by comparing their predictions with the ground truth data obtained from inertial measurement units (IMUs), which were considered as the ground truth for out-of-the-lab and non-clinic settings. The evaluation was conducted using several common and robust error metrics, including Root Mean Squared Error (RMSE), Mean Absolute Error (MAE), Normalized RMSE (NRMSE), Pearson's correlation coefficient, and the coefficient of determination ($R^2$).

We selected RMSE and MAE to quantify absolute angular error in degrees because they provide complementary sensitivity to outliers and are directly interpretable for clinical kinematics. We report NRMSE to enable fair comparisons across joints and activities that have different motion ranges. Pearson correlation is included to capture temporal agreement between the filtered and synchronized angle trajectories. $R^2$ summarizes the proportion of variance explained by the model at the waveform level. Since our evaluation is performed on joint angles rather than 3D joint coordinates, coordinate based pose metrics such as MPJPE or PCK are not used here. The following sections detail the evaluation metrics and statistical methods used in this study:

1) **Root Mean Squared Error (RMSE):** RMSE was used as the primary error metric to quantify the overall accuracy of the pose estimations. It calculates the square root of the mean squared differences between the predicted joint angles and the ground truth from IMU data. Therefore, it penalizes larger errors more due to the squared elevation, which makes it more sensitive to outliers. RMSE is a widely used metric due to its sensitivity to large errors and its interpretation in the same units as the data (degrees for angular measurements). It is computed following the expression in Eq. (2).

$$RMSE = \sqrt{\frac{1}{N}\sum_{i=1}^{N}(y_i - \hat{y}_i)^2} \qquad (2)$$

Where $y_i$ are the ground truth joint angles, $\hat{y}_i$ are the predicted joint angles, and $N$ is the total number of samples. Lower RMSE values indicate better accuracy.

2) **Normalized Root Mean Square Error (NRMSE):** To account for the range of joint angles, NRMSE was computed by normalizing the RMSE by the range (or standard deviation) of the ground truth signal (IMUs signal), as shown in Eq. (3). This normalization allows for the comparison of RMSE values across different poses and activities with varying ranges of human body three-dimensional joint motion.

$$NRMSE = \frac{RMSE}{\text{Range}(y)} \qquad (3)$$

Where $Range(y)$ is the difference between the maximum and minimum values of the ground truth signal.

3) **Mean Absolute Error (MAE):** To provide a more intuitive measure of prediction accuracy, MAE was also computed following the definition in Eq. (4). This metric represents the average of the absolute differences between the estimated and measured values. Unlike RMSE, MAE treats all errors equally (all errors have the same weight) and provides a direct average of the absolute errors, being more robust to outlier errors.

$$MAE = \frac{1}{N}\sum_{i=1}^{N}|y_i - \hat{y}_i| \qquad (4)$$

Where $y_i$ are the ground truth values, $\hat{y}_i$ are the estimated values, and $N$ is the total number of samples.

4) **Pearson's Correlation Coefficient:** The correlation coefficient in Eq. (5) was used to measure the linear relationship between the predicted joint angles and the ground truth angles extracted from the IMUs. A value close to 1 indicates a strong positive correlation, while a value close to -1 indicates a strong negative correlation. A value near 0 indicates no linear relationship. This metric quantifies the strength of the linear relationship between the IMU measurements and the video-based estimates.

$$\text{Correlation} = \frac{\sum_{i=1}^{N}(y_i - \bar{y})(\hat{y}_i - \bar{\hat{y}})}{\sqrt{\sum_{i=1}^{N}(y_i - \bar{y})^2 \sum_{i=1}^{N}(\hat{y}_i - \bar{\hat{y}})^2}} \qquad (5)$$

Where $\bar{y}$ and $\bar{\hat{y}}$ are the mean values of the IMU- and video-based signals, respectively.

5) **Coefficient of Determination ($R^2$):** The Coefficient of Determination was used to assess the proportion of the variance in the ground truth data (dependent variable) that is predictable or explained by the model data (independent variable(s)). Its expression is defined in Eq. (6). A high value (close to 1) indicates that the model explains a high proportion of the variance in the data, which is expected for a model that fits the data correctly. An $R^2$ value close to zero implies that the model does not explain any of the variability of the response data around its mean. In such a case, the model's predictions are no better than simply using the mean of the observed data as a prediction for every observation. Negative values indicate that the model performs worse than simply predicting the mean.

$$R^2 = 1 - \frac{\sum_{i=1}^{N}(y_i - \widehat{y}_i)^2}{\sum_{i=1}^{N}(y_i - \overline{y})^2} \qquad (6)$$

All the above metrics were computed for each subject and activity, as well as for the entire dataset, allowing us to compare the overall performance of each deep learning HPE model. Our evaluation pipeline involved processing the raw video-based and IMU signals before computing the metrics. For each activity, per–subject metrics were aggregated to produce summary tables and bar charts for RMSE, MAE, Pearson correlation, and $R^2$. In addition, the overall performance of each model was assessed by averaging these metrics across all subjects and activities. The aggregated summary table reports the mean and standard deviation of each metric per model, thus facilitating a direct comparison of their overall performance. All plots and tables were saved in both SVG and PDF formats, ensuring high-quality figures for inclusion in the manuscript.

This comprehensive set of evaluation metrics and the corresponding statistical analyses enable a robust assessment of the deep-learning models' performance. They provide a basis for discussing trade-offs between accuracy (e.g., lower RMSE and MAE), agreement, and consistency (e.g., high Pearson correlation), ultimately supporting our conclusion regarding which model performs best overall. With the data, models, and metrics defined, we now present the results of the benchmark.

## 3. Results

The performance of the four video-based 3D HPE systems (MotionAGFormer (Mehraban et al., 2023), MotionBERT (Zhu et al., 2022), MMPose, and NVIDIA BodyTrack) was evaluated against IMU-derived joint angles across a range of daily-life upper- and lower-limb activities. Error metrics including Root Mean Squared Error (RMSE), Mean Absolute Error (MAE), Normalized RMSE (NRMSE), Pearson's correlation coefficient, and the coefficient of determination ($R^2$) were computed for each method, both on an overall basis and for individual tasks. Thus, the performance of the models was assessed both holistically and with a specific focus on activities targeting distinct regions of the human body. We report results at two levels. We first summarize overall performance across all activities and then examine activity specific patterns.

### 3.1. Overall performance

The overall results revealed significant variations in performance across models and tasks, as well as distinct trade-offs. Aggregated results across all 13 activities, which are illustrated in Fig. **4** and further detailed in Table **3**, demonstrated that MotionAGFormer achieved the lowest overall RMSE (9.27° ± 4.80°) and MAE (7.86° ± 4.18°), as well as the highest correlation (0.86 ± 0.15) and the highest coefficient of determination $R^2$ (0.67 ± 0.28), indicating superior consistency in approximating IMU measurements compared to the rest of the models. In contrast, MotionBERT registered the highest errors for the overall performance, with the highest RMSE (12.28° ± 4.59°) and the highest MAE (10.15° ± 3.86°), coupled with the second lowest correlation coefficient (0.79 ± 0.11), and the lowest $R^2$ (0.16 ± 0.50). BodyTrack and MMPose produced intermediate results, with mean RMSE values of 10.89° ± 3.67° and 11.05° ± 4.17°, MAE values of 9.00 ± 3.12 and 9.35 ± 3.61, correlation values of 0.78 ± 0.12 and 0.84 ± 0.10, and $R^2$ values of 0.44 ± 0.31 and 0.58 ± 0.26, respectively.

### 3.2. Activity-specific performance

The evaluation was further stratified by activity, revealing variations in model performance according to the nature of the movement, the occlusions, and the human body parts involved. Detailed metrics for the RMSE and the correlation coefficient can be found in Table **4** and Table **5** respectively. Within each task, evaluation metrics were also calculated individually for each participant in the dataset, as shown in Fig. **5**. Tables with the rest of the metrics and plots for each activity are included in the Supplementary Material section. All generated graphs and tables are available in Zenodo (Medrano-Paredes et al., 2025).

For Lower Limb Activities, and more specifically for walking-related tasks, the error metrics were generally lower. In Activity A01 (walk_forward), MMPose yielded the best performance with an RMSE of 6.54° ± 2.64° and an $R^2$ of 0.76 ± 0.20. Activity A02 (walk_backward) had NVIDIA BodyTrack leading with an RMSE of 4.72° ± 1.40° and an $R^2$ of 0.80 ± 0.10, while in Activity A03 (walk_along), MotionAGFormer recorded an RMSE of 5.66° ± 2.72° and an $R^2$ of 0.83 ± 0.12. For the last lower limb task, Activity A04 (sit_to_stand), MotionBERT showed the best results, with an RMSE of 5.71 ± 2.10 and an $R^2$ of 0.95 ± 0.04.

For Upper Limb Activities, involving more complex and less predictable movements, the performance differences across different models became more pronounced. For example, in activity A08 (drink_left_arm), NVIDIA BodyTrack and MMPose achieved the best coefficient of determination with very close values of $R^2$ 0.82 ± 0.11 for both models. In activity A09 (assemble_both_arms), MMPose remained the leading model with an $R^2$ of 0.72 ± 0.18, followed by MotionAGFormer with an $R^2$ of 0.67 ± 0.29, the rest of the models being much less accurate. For activity A12 (reachup_left_arm), MotionBERT performed best with an $R^2$ of 0.98 ± 0.02, followed by MMPose and BodyTrack with similar values of $R^2$ around 0.95 ± 0.03. Finally, in activity A13 (tear_both_arms), MotionAGFormer exhibited the best accuracy with an $R^2$ of 0.94 ± 0.06, followed by BodyTrack with an $R^2$ of 0.87 ± 0.11. The corresponding bar charts and metrics tables consistently showed that MMPose and MotionAGFormer maintained moderate error levels and robust correlations across subjects. Conversely, MotionBERT frequently exhibited increased RMSE and MAE values along with reduced Pearson correlation and determination coefficients, suggesting reduced robustness when capturing the finer kinematics of upper limb tasks.

These results underscore the viability of video-based 3D HPE as a scalable alternative to IMUs, particularly for applications prioritizing accessibility and cost-efficiency. While all evaluated deep learning methods are viable for estimating 3D human pose in unconstrained, daily-life settings, the results highlight clear trade-offs. Thus, the choice of model must align with specific clinical requirements, considering that some of the systems showed the best performance for lower limb detection tasks while others achieved the best results in more general, holistic activities. With the aggregate and activity specific findings in view, it is crucial to interpret these results, discuss model behavior by task, and delimit the scope of our claims.

## 4. Discussion and limitations

### 4.1. Discussion

This subsection interprets the comparative results in light of model design and data characteristics, and links them to intended uses in out-of-the-lab kinematic assessment. The comparative analysis reveals that each 3D HPE model exhibits distinct strengths and limitations, which appear to be closely tied to their underlying architectures, training paradigms, and intended applications. These differences become particularly evident when comparing overall aggregated performance metrics as well as activity-specific results.

MotionAGFormer emerged as the top performer overall (e.g. RMSE: 9.27° ± 4.80° and $R^2$: 0.67 ± 0.28), consistently achieving the lowest RMSE and MAE, and the highest correlation and $R^2$ across several activities. Its dual-stream hybrid architecture, which integrates transformer-based self-attention (Vaswani et al., 2023) for global spatio-temporal dependencies with Graph Convolutional Networks (GCNs) (Kipf & Welling, 2017) for capturing local joint relationships, appears to enable adaptive fusion of critical features for both detecting complex movements and capturing fine-grained joint dependencies more effectively. For instance, during bimanual activities, its dual-stream design efficiently resolves occlusions and rapid motion changes, while sequence-level predictions ensure temporal smoothness. This design not only leverages the strengths of both transformer and GCN modules but also provides a more balanced representation of spatio-temporal dynamics, which is critical in unconstrained, daily-life scenarios. Note that, although MotionAGFormer achieves the highest accuracy, it also has the longest inference time among the evaluated models., which may be a disadvantage in situations where fast evaluations are needed or even in future developments oriented towards real-time pose detection.

In contrast, MotionBERT demonstrates a more mixed performance profile, where a reduced inference time can be emphasized respect to the rest of the models. While its pretraining strategy, focused on learning robust motion representations from corrupted 2D inputs, holds promise for a range of downstream tasks, the aggregated metrics indicate higher errors (e.g. overall RMSE: 12.28° ± 4.59° and overall $R^2$: 0.16 ± 0.50) due to its reliance on self-supervised pretraining. For instance, despite strong performance in certain lower-limb tasks, as seen in Activity A04 (sit_to_stand), with an RMSE of 5.71° ± 2.10° and an $R^2$ of 0.95 ± 0.04, its performance deteriorates in other scenarios, suggesting that the learned motion priors may not generalize uniformly across all types of movements. While MotionBERT's pretraining on corrupted skeletons enhances robustness to noise and occlusions, it may inadequately generalize to fine-grained joint kinematics in dynamic tasks, particularly where precise depth estimation is required. Also, although the pretraining approach allows it to incorporate geometric and kinematic priors, its performance may be limited by the challenges of transferring these learned representations to the specific conditions of the VIDIMU dataset. Additionally, while its design reflects an innovative unified framework for human motion understanding, its architectural choices and earlier release date relative to some of the newer models might have constrained its ability to capture the subtle dynamics required for precise joint angle estimation.

MMPose demonstrated strong performance in lower-limb tasks such as Activity A01 (walk_forward), with an RMSE of 6.54° ± 2.64° and an $R^2$ value of 0.76 ± 0.20, while showing intermediate results in the rest of the tasks. Its modular three-stage pipeline (detection, 2D pose estimation, 2D-to-3D pose lifting) benefits from domain-specific pretraining on datasets like Human3.6M (Ionescu et al., 2014), which emphasizes gait analysis. However, its performance degraded in upper-limb tasks (e.g. in Activity A13 tear_both_arms, with RMSE: 17.57° ± 3.53° and $R^2$: 0.34 ± 0.40), likely due to limited temporal window size in VideoPoseLift (243 frames), which struggles with prolonged, non-repetitive motions. This is likely due to the fact that the reliance on a sequential process introduces potential cumulative errors, particularly in activities with significant occlusions or rapid movements. However, its competitive performance metrics in some tasks suggest that, in scenarios where 2D detections are reliable, the lifting approach can provide acceptable estimates of 3D poses.

The latest model included in the benchmark is NVIDIA BodyTrack. While being a proprietary solution and opaque in design, it achieved competitive results across both lower- and upper-limb activities. However, its higher error metrics in occluded scenarios (e.g. in Activity A09 assemble_both_arms with an RMSE of 10.15° ± 4.00° and an $R^2$ of 0.19 ± 0.38) highlight a strong reliance on controlled non-occluded environments. Although BodyTrack is optimized for real-time applications and more features other than HPE, its lack of transparency and potential limitations in adapting to diverse environments may hinder its performance in settings that differ from those originally targeted by the developer. The interoperability is also affected by the fact that BodyTrack outputs 34 human body markers, while the rest of the models output a total of 17 markers, following the format of Human3.6M dataset.

Several factors contribute to these performance differences. The architectures of newer models like MotionAGFormer (2023) reflect recent advancements in Deep Learning, where hybrid designs offer improved trade-offs between accuracy and computational efficiency. The release often correlates with such advancements, as models released more recently have benefited from larger, more diverse datasets and refined training methodologies that address known limitations in earlier approaches, especially in fields such as Deep Learning and Computer Vision where scientific progress is advancing at a rapid pace. On the other hand, models developed earlier, such as MMPose (2022), MotionBERT (2022) or NVIDIA BodyTrack (2021), may lack these optimizations, which is evident in the performance gap observed in this study.

It is also worthwhile to highlight the use cases and the most suitable applications for each one. MotionAGFormer, with its hybrid transformer-GCN architecture, excels in complex and dynamic lower- and upper-limb activities. This makes it ideal for applications requiring high precision, such as advanced telerehabilitation and sports performance analysis, where capturing both global context and fine-grained joint interactions is critical. In contrast, MotionBERT shines in predictable, structured and linear trajectories due to its self-supervised pretraining on diverse motion data, but it struggles with abrupt movements and positioning. However, its slightly higher error margins may limit its use in applications where exact joint angle accuracy is paramount. Meanwhile, MMPose, leveraging a sequential 2D-to-3D lifting approach, provides a balanced solution when high-quality 2D keypoint detections are available and slight compromises in precision are acceptable. Therefore, it represents a modular, cost-effective solution for applications prioritizing lower-limb and gait-focused analysis. NVIDIA BodyTrack, optimized for a balanced performance and integrated within AR/VR platforms is best suited for scenarios prioritizing speed and operational simplicity, despite its closed-source nature restricting adaptability as well as further academic scrutiny and optimization. Considering that no single model universally outperforms others, the results underscore the importance of selecting an appropriate model based on the specific application and activity type.

We next summarize the main limitations that qualify these interpretations and indicate where careful consideration is advised.

### 4.2. Limitations

Limitations of this preclinical benchmark must also be acknowledged. The VIDIMU dataset (Martínez-Zarzuela et al., 2023), although designed to replicate daily-life activities, involves a limited number of subjects, tasks, and controlled conditions that might not fully capture the variability, diverse environments, and strong occlusions encountered in real-world scenarios. For example, varying lighting conditions (S. Lee et al., 2023), camera perspectives and location, physical constraints (Hsu & Jang, 2024), and occlusion patterns (Lino et al., 2024) need to be addressed in the future. Each model's performance is also influenced by its temporal resolution (e.g., frame-by-frame predictions lack long-term coherence, whereas temporal windowing better handles prolonged activities) as well as by its specific pretraining or fine-tuning regimen, which might not be fully optimal for the diverse set of activities included in our evaluation. On the other hand, hardware constraints should also be considered: IMU-based ground truth is inherently influenced by the accuracy of the IMU-to-segment calibration performed during VIDIMU acquisition for every activity recording, potentially biasing error metrics and introducing a systematic source of error. In the case of NVIDIA BodyTrack, the debate arises between proprietary vs. open-source AI models, as the dependency on NVIDIA's ecosystem limits reproducibility and modularity, while open-source models offer data transparency and flexibility but require significant computational resources for training. Also, its closed-source nature limits modularity and raises some concerns about privacy when evaluating patient data in clinic (Naresh et al., 2023) or industry applications. While IMUs are well suited for remote use and serve as a practical out-of-the-lab reference in this study, optical motion capture remains the in-lab standard for joint-angle validation. A follow-up study should include an optical system such as Vicon when available in order to triangulate estimates across modalities and report laboratory-grade agreement metrics.

This preclinical study includes healthy adults only, which limits direct generalization to clinical populations. Patients with neurological or musculoskeletal conditions often present altered movement patterns, use of assistive devices and greater general variability, all of which can change the behavior of video-based estimators and IMU-derived references. Working with a single, uniform dataset improves internal validity and fairness across models, yet it restricts external validity. We acknowledge the absence of pathological participants and note that adding patient data without a matched protocol could introduce confounding and bias the comparative analysis.

With these constraints in mind, we close with the key takeaways and outline the next steps toward clinical validation and robust application.

## 5. Conclusion and future work

### 5.1. Conclusion

This subsection synthesizes the main results into concise guidance for model choice and deployment under out-of-the-lab conditions. The study presents a comprehensive benchmark of four state-of-the-art video-based monocular 3D human pose estimation (HPE) models—MotionAGFormer, MotionBERT, MMPose 2D-to-3D lifting pipeline, and NVIDIA BodyTrack—against inertial measurement units IMU-derived joint angles, aiming to evaluate their feasibility for clinical telemonitoring and remote rehabilitation. Our analysis, grounded in robust error metrics (RMSE, MAE, NRMSE, Pearson's correlation coefficient, and coefficient of determination $R^2$), reveals that each model exhibits unique trade-offs between accuracy, inference time, computational efficiency and adaptability. Notably, MotionAGFormer consistently achieved superior overall performance by effectively balancing global and local spatio-temporal dependencies thanks to its dual hybrid architecture integrating Transformers and Graph Convolutional Networks (GCNs), being the most accurate model for the evaluated dataset.

These findings underscore the potential of video-based 3D HPE as a viable, non-intrusive, cost-effective alternative to IMUs and other optical- and sensor-based motion capture solutions, with wide applications in the fields of telerehabilitation, sports performance analysis, and remote patient monitoring. The choice of model should therefore be guided by the specific application context—whether prioritizing precision in dynamic movements, computing time, or operational simplicity in resource-constrained settings. Our work thereby bridges the precision-accessibility gap by providing critical insights into the strengths and limitations of current video-based HPE models, laying the groundwork for future advancements that could revolutionize the monitoring and analysis of human kinematics outside typical laboratory settings. Building on these findings, the next subsection details specific avenues to broaden validity, enhance robustness, and enable practical use.

### 5.2. Future work

To further democratize out-of-the-lab motion capture, future research should prioritize hybrid systems that synergize video and sensor modalities (Marcard et al., 2016), leveraging the strengths of both and enhancing robustness and accuracy in real-world scenarios. Expanding evaluations to diverse environmental conditions (e.g., lighting (S. Lee et al., 2023), camera position and orientation, physical constraints (Hsu & Jang, 2024), occlusions (Lino et al., 2024), multiple subjects in the frame (T. Jiang et al., 2024; Luo et al., 2010)), will enhance clinical relevance. Domain-specific fine-tuning of models (Joo et al., 2021), particularly for underrepresented upper-limb tasks, could improve robustness. Innovations in temporal modeling—such as adaptive windowing or attention mechanisms—may address current limitations in capturing prolonged or non-repetitive motions.

To improve generalization in clinical settings, future studies should evaluate fine-tuning and domain adaptation strategies tailored to 3D HPE. First, supervised fine-tuning on a small set of target recordings collected with the same camera placement and protocol can calibrate models to the clinical domain while controlling labeling cost. Second, self-supervised adaptation that exploits large unlabeled videos from the clinic through consistency, temporal assembling, or pseudo labels can reduce reliance on annotations. Third, realistic data augmentation and domain randomization that vary lighting, viewpoint, background, occlusion, and clothing can be applied to RGB frames before lifting. On the other hand, parameter-efficient techniques allow specialization with limited compute, which is attractive for edge devices. Cross-modal distillation that uses synchronized IMU signals to guide video models can align the two modalities in out-of-the-lab conditions. Together, these strategies can turn the off-the-shelf baselines reported here into deployable systems that are resilient to clinical variability while preserving reproducibility.

One of the main priorities is to build an extended multimodal dataset using the same setup and protocol (monocular video and IMUs recorded in the semi-sagittal plane at 45º), but now with clinically recruited participants alongside healthy controls. Using the identical setup and processing pipeline will preserve comparability and allow a fair extension of the benchmark to real clinical scenarios. This resource will facilitate the development of mappings between continuous kinematic features and clinician-rated instruments.

Finally, future studies should explicitly connect objective kinematic measures with established clinical outcome scales such as UPDRS and the Berg Balance Scale. This will involve recording the scale specific tasks under the same single camera and IMU protocol, recruiting appropriate clinical cohorts, and designing feature extraction and modeling pipelines that link joint angle signals to discrete clinical scores while accounting for inter rater variability, reliability, and minimal detectable change. The goal is to establish compatibility between metric based, continuous kinematic assessments and clinician-rated instruments so that video-based methods can complement routine evaluation and support decision making in real-world practice.

**References**

Andriluka, M., Iqbal, U., Insafutdinov, E., Pishchulin, L., Milan, A., Gall, J., & Schiele, B. (2018). *PoseTrack: A Benchmark for Human Pose Estimation and Tracking* (No. arXiv:1710.10000). arXiv. https://doi.org/10.48550/arXiv.1710.10000

Antón, D., Goñi, A., Illarramendi, A., Torres-Unda, J. J., & Seco, J. (2013). KiReS: A Kinect-based telerehabilitation system. *2013 IEEE 15th International Conference on E-Health Networking, Applications and Services (Healthcom 2013)*, 444–448. https://doi.org/10.1109/HealthCom.2013.6720717

Avogaro, A., Cunico, F., Rosenhahn, B., & Setti, F. (2023). Markerless human pose estimation for biomedical applications: A survey. *Frontiers in Computer Science*, *5*. https://doi.org/10.3389/fcomp.2023.1153160

Badiola-Bengoa, A., & Mendez-Zorrilla, A. (2021). A Systematic Review of the Application of Camera-Based Human Pose Estimation in the Field of Sport and Physical Exercise. *Sensors*, *21*(18), Article 18. https://doi.org/10.3390/s21185996

Ben Gamra, M., & Akhloufi, M. A. (2021). A review of deep learning techniques for 2D and 3D human pose estimation. *Image and Vision Computing*, *114*, 104282. https://doi.org/10.1016/j.imavis.2021.104282

Chakraborty, P., & Namboodiri, V. P. (2017). *Learning to Estimate Pose by Watching Videos* (No. arXiv:1704.04081). arXiv. https://doi.org/10.48550/arXiv.1704.04081

Chen, K., Gabriel, P., Alasfour, A., Gong, C., Doyle, W. K., Devinsky, O., Friedman, D., Dugan, P., Melloni, L., Thesen, T., Gonda, D., Sattar, S., Wang, S., & Gilja, V. (2018). Patient-Specific Pose Estimation in Clinical Environments. *IEEE Journal of Translational Engineering in Health and Medicine*, *6*, 1–11. IEEE Journal of Translational Engineering in Health and Medicine. https://doi.org/10.1109/JTEHM.2018.2875464

Chen, Y., Tian, Y., & He, M. (2020). Monocular human pose estimation: A survey of deep learning-based methods. *Computer Vision and Image Understanding*, *192*, 102897. https://doi.org/10.1016/j.cviu.2019.102897

Fukushima, K. (1980). Neocognitron: A self-organizing neural network model for a mechanism of pattern recognition unaffected by shift in position. *Biological Cybernetics*, *36*(4), 193–202. https://doi.org/10.1007/BF00344251

Höglund, G., Grip, H., & Öhberg, F. (2021). The importance of inertial measurement unit placement in assessing upper limb motion. *Medical Engineering & Physics*, *92*, 1–9. https://doi.org/10.1016/j.medengphy.2021.03.010

Hsu, C., & Jang, J. (2024). *BLAPose: Enhancing 3D Human Pose Estimation with Bone Length Adjustment* (No. arXiv:2410.20731; Version 1). arXiv. https://doi.org/10.48550/arXiv.2410.20731

Huang, Z., Liu, Y., Fang, Y., & Horn, B. K. P. (2018). Video-based Fall Detection for Seniors with Human Pose Estimation. *2018 4th International Conference on Universal Village (UV)*, 1–4. https://doi.org/10.1109/UV.2018.8642130

Ienaga, N., Takahata, S., Terayama, K., Enomoto, D., Ishihara, H., Noda, H., & Hagihara, H. (2022). Development and Verification of Postural Control Assessment Using Deep-Learning-Based Pose Estimators: Towards Clinical Applications. *Occupational Therapy International*, *2022*(1), 6952999. https://doi.org/10.1155/2022/6952999

Insafutdinov, E., Pishchulin, L., Andres, B., Andriluka, M., & Schiele, B. (2016). *DeeperCut: A Deeper, Stronger, and Faster Multi-Person Pose Estimation Model* (No. arXiv:1605.03170). arXiv. https://doi.org/10.48550/arXiv.1605.03170

Ionescu, C., Papava, D., Olaru, V., & Sminchisescu, C. (2014). Human3.6M: Large Scale Datasets and Predictive Methods for 3D Human Sensing in Natural Environments. *IEEE Transactions on Pattern Analysis and Machine Intelligence*, *36*(7), 1325–1339. https://doi.org/10.1109/TPAMI.2013.248

Jiang, B., Xu, F., Zhang, Z., Tang, J., & Nie, F. (2023). *AGFormer: Efficient Graph Representation with Anchor-Graph Transformer* (No. arXiv:2305.07521). arXiv. https://doi.org/10.48550/arXiv.2305.07521

Jiang, T., Lu, P., Zhang, L., Ma, N., Han, R., Lyu, C., Li, Y., & Chen, K. (2023). *RTMPose: Real-Time Multi-Person Pose Estimation based on MMPose* (No. arXiv:2303.07399). arXiv. https://doi.org/10.48550/arXiv.2303.07399

Jiang, T., Xie, X., & Li, Y. (2024). *RTMW: Real-Time Multi-Person 2D and 3D Whole-body Pose Estimation* (No. arXiv:2407.08634). arXiv. https://doi.org/10.48550/arXiv.2407.08634

Jiang, Z., Zhou, Z., Li, L., Chai, W., Yang, C.-Y., & Hwang, J.-N. (2023). *Back to Optimization: Diffusion-based Zero-Shot 3D Human Pose Estimation* (No. arXiv:2307.03833). arXiv. https://doi.org/10.48550/arXiv.2307.03833

Joo, H., Neverova, N., & Vedaldi, A. (2021). Exemplar Fine-Tuning for 3D Human Model Fitting Towards In-the-Wild 3D Human Pose Estimation. *2021 International Conference on 3D Vision (3DV)*, 42–52. https://doi.org/10.1109/3DV53792.2021.00015

Kipf, T. N., & Welling, M. (2017). *Semi-Supervised Classification with Graph Convolutional Networks* (No. arXiv:1609.02907). arXiv. https://doi.org/10.48550/arXiv.1609.02907

Lea, C., Vidal, R., Reiter, A., & Hager, G. D. (2016). *Temporal Convolutional Networks: A Unified Approach to Action Segmentation* (No. arXiv:1608.08242). arXiv. https://doi.org/10.48550/arXiv.1608.08242

Lee, S., Rim, J., Jeong, B., Kim, G., Woo, B., Lee, H., Cho, S., & Kwak, S. (2023). *Human Pose Estimation in Extremely Low-Light Conditions* (No. arXiv:2303.15410). arXiv. https://doi.org/10.48550/arXiv.2303.15410

Lee, Y., Lama, B., Joo, S., & Kwon, J. (2024). Enhancing Human Key Point Identification: A Comparative Study of the High-Resolution VICON Dataset and COCO Dataset Using BPNET. *Applied Sciences*, *14*(11), Article 11. https://doi.org/10.3390/app14114351

Lin, T.-Y., Dollár, P., Girshick, R., He, K., Hariharan, B., & Belongie, S. (2017). *Feature Pyramid Networks for Object Detection* (No. arXiv:1612.03144). arXiv. https://doi.org/10.48550/arXiv.1612.03144

Lin, T.-Y., Maire, M., Belongie, S., Bourdev, L., Girshick, R., Hays, J., Perona, P., Ramanan, D., Zitnick, C. L., & Dollár, P. (2015). *Microsoft COCO: Common Objects in Context* (No. arXiv:1405.0312). arXiv. https://doi.org/10.48550/arXiv.1405.0312

Lino, F., Santiago, C., & Marques, M. (2024). *3D Human Pose Estimation with Occlusions: Introducing BlendMimic3D Dataset and GCN Refinement* (No. arXiv:2404.16136). arXiv. https://doi.org/10.48550/arXiv.2404.16136

Luo, X., Berendsen, B., Tan, R. T., & Veltkamp, R. C. (2010). Human Pose Estimation for Multiple Persons Based on Volume Reconstruction. *2010 20th International Conference on Pattern Recognition*, 3591–3594. https://doi.org/10.1109/ICPR.2010.876

Lyu, C., Zhang, W., Huang, H., Zhou, Y., Wang, Y., Liu, Y., Zhang, S., & Chen, K. (2022). *RTMDet: An Empirical Study of Designing Real-Time Object Detectors* (No. arXiv:2212.07784). arXiv. https://doi.org/10.48550/arXiv.2212.07784

Marcard, T. von, Pons-Moll, G., & Rosenhahn, B. (2016). Human Pose Estimation from Video and IMUs. *IEEE Transactions on Pattern Analysis and Machine Intelligence*, *38*(8), 1533–1547. IEEE Transactions on Pattern Analysis and Machine Intelligence. https://doi.org/10.1109/TPAMI.2016.2522398

Martínez-Zarzuela, M., González-Alonso, J., Antón-Rodríguez, M., Díaz-Pernas, F. J., Müller, H., & Simón-Martínez, C. (2023). Multimodal video and IMU kinematic dataset on daily life activities using affordable devices. *Scientific Data*, *10*(1), 648. https://doi.org/10.1038/s41597-023-02554-9

Martini, E., Boldo, M., Aldegheri, S., Valè, N., Filippetti, M., Smania, N., Bertucco, M., Picelli, A., & Bombieri, N. (2022). Enabling Gait Analysis in the Telemedicine Practice through Portable and Accurate 3D Human Pose Estimation. *Computer Methods and Programs in Biomedicine*, *225*, 107016. https://doi.org/10.1016/j.cmpb.2022.107016

Medrano-Paredes, M., Saoudi, H., Fernández-González, C., Díaz-Pernas, F. J., González-Ortega, D., Antón-Rodríguez, M., & Martínez-Zarzuela, M. (2025). *Paving the Way Towards Kinematic Assessment Using Monocular Video: A Benchmark of State-of-the-Art Deep-Learning-Based 3D Human Pose Estimators Against Inertial Sensors in Daily Living Activities* (Version 1.0.0) [Dataset]. Zenodo. https://doi.org/10.5281/zenodo.15088423

Mehraban, S., Adeli, V., & Taati, B. (2023). *MotionAGFormer: Enhancing 3D Human Pose Estimation with a Transformer-GCNFormer Network* (No. arXiv:2310.16288). arXiv. https://doi.org/10.48550/arXiv.2310.16288

Mehta, D., Rhodin, H., Casas, D., Fua, P., Sotnychenko, O., Xu, W., & Theobalt, C. (2017). *Monocular 3D Human Pose Estimation In The Wild Using Improved CNN Supervision* (No. arXiv:1611.09813). arXiv. https://doi.org/10.48550/arXiv.1611.09813

Milosevic, B., Leardini, A., & Farella, E. (2020). Kinect and wearable inertial sensors for motor rehabilitation programs at home: State of the art and an experimental comparison. *BioMedical Engineering OnLine*, *19*(1), 25. https://doi.org/10.1186/s12938-020-00762-7

Mizuochi, Y., Shigematsu, Y., & Fukuura, Y. (2024). Recovery environments in places of daily living: A scoping review and conceptual analysis. *BMC Public Health*, *24*(1), 3046. https://doi.org/10.1186/s12889-024-20489-7

Naresh, V. S., Thamarai, M., & Allavarpu, V. V. L. D. (2023). Privacy-preserving deep learning in medical informatics: Applications, challenges, and solutions. *Artificial Intelligence Review*, *56*(1), 1199–1241. https://doi.org/10.1007/s10462-023-10556-7

Nijmeijer, E. M., Heuvelmans, P., Bolt, R., Gokeler, A., Otten, E., & Benjaminse, A. (2023). Concurrent validation of the Xsens IMU system of lower-body kinematics in jump-landing and change-of-direction tasks. *Journal of Biomechanics*, *154*, 111637. https://doi.org/10.1016/j.jbiomech.2023.111637

Niswander, W., Wang, W., & Kontson, K. (2020). Optimization of IMU Sensor Placement for the Measurement of Lower Limb Joint Kinematics. *Sensors*, *20*(21), Article 21. https://doi.org/10.3390/s20215993

Ouyang, W., Chu, X., & Wang, X. (2014). Multi-source Deep Learning for Human Pose Estimation. *2014 IEEE Conference on Computer Vision and Pattern Recognition*, 2337–2344. https://doi.org/10.1109/CVPR.2014.299

Pang, H. E., Cai, Z., Yang, L., Zhang, T., & Liu, Z. (2022). *Benchmarking and Analyzing 3D Human Pose and Shape Estimation Beyond Algorithms* (No. arXiv:2209.10529). arXiv. https://doi.org/10.48550/arXiv.2209.10529

Poitras, I., Dupuis, F., Bielmann, M., Campeau-Lecours, A., Mercier, C., Bouyer, L. J., & Roy, J.-S. (2019). Validity and Reliability of Wearable Sensors for Joint Angle Estimation: A Systematic Review. *Sensors*, *19*(7), Article 7. https://doi.org/10.3390/s19071555

Rosique, F., Losilla, F., & Navarro, P. J. (2021). Applying Vision-Based Pose Estimation in a Telerehabilitation Application. *Applied Sciences*, *11*(19), Article 19. https://doi.org/10.3390/app11199132

Seidman, Z., McNamara, R., Wootton, S., Leung, R., Spencer, L., Dale, M., Dennis, S., & McKeough, Z. (2017). People attending pulmonary rehabilitation demonstrate a substantial engagement with technology and willingness to use telerehabilitation: A survey. *Journal of Physiotherapy*, *63*(3), 175–181. https://doi.org/10.1016/j.jphys.2017.05.010

Topham, L. K., Khan, W., Al-Jumeily, D., & Hussain, A. (2023). Human Body Pose Estimation for Gait Identification: A Comprehensive Survey of Datasets and Models. *ACM Computing Surveys*, *55*(6), 1–42. https://doi.org/10.1145/3533384

Toshev, A., & Szegedy, C. (2014). DeepPose: Human Pose Estimation via Deep Neural Networks. *2014 IEEE Conference on Computer Vision and Pattern Recognition*, 1653–1660. https://doi.org/10.1109/CVPR.2014.214

Vafadar, S., Skalli, W., Bonnet-Lebrun, A., Assi, A., & Gajny, L. (2022). Assessment of a novel deep learning-based marker-less motion capture system for gait study. *Gait & Posture*, *94*, 138–143. https://doi.org/10.1016/j.gaitpost.2022.03.008

Vaswani, A., Shazeer, N., Parmar, N., Uszkoreit, J., Jones, L., Gomez, A. N., Kaiser, L., & Polosukhin, I. (2023). *Attention Is All You Need* (No. arXiv:1706.03762). arXiv. https://doi.org/10.48550/arXiv.1706.03762

Von Marcard, T., Henschel, R., Black, M. J., Rosenhahn, B., & Pons-Moll, G. (2018). Recovering Accurate 3D Human Pose in the Wild Using IMUs and a Moving Camera. In V. Ferrari, M. Hebert, C. Sminchisescu, & Y. Weiss (Eds.), *Computer Vision – ECCV 2018* (Vol. 11214, pp. 614–631). Springer International Publishing. https://doi.org/10.1007/978-3-030-01249-6_37

Wang, C.-Y., Liao, H.-Y. M., Yeh, I.-H., Wu, Y.-H., Chen, P.-Y., & Hsieh, J.-W. (2019). *CSPNet: A New Backbone that can Enhance Learning Capability of CNN* (No. arXiv:1911.11929). arXiv. https://doi.org/10.48550/arXiv.1911.11929

Wang, J., Sun, K., Cheng, T., Jiang, B., Deng, C., Zhao, Y., Liu, D., Mu, Y., Tan, M., Wang, X., Liu, W., & Xiao, B. (2020). *Deep High-Resolution Representation Learning for Visual Recognition* (No. arXiv:1908.07919). arXiv. https://doi.org/10.48550/arXiv.1908.07919

Wang, J., Tan, S., Zhen, X., Xu, S., Zheng, F., He, Z., & Shao, L. (2021). Deep 3D human pose estimation: A review. *Computer Vision and Image Understanding*, *210*, 103225. https://doi.org/10.1016/J.CVIU.2021.103225

Wilhelm, N., Micheler, C. M., Lang, J. J., Hinterwimmer, F., Schaack, V., Smits, R., Haddadin, S., & Burgkart, R. (2023). Development and Evaluation of a Cost-effective IMU System for Gait Analysis: Comparison with Vicon and VideoPose3D Algorithms. *Current Directions in Biomedical Engineering*, *9*(1), 254–257. https://doi.org/10.1515/cdbme-2023-1064

Wittmann, F., Lambercy, O., & Gassert, R. (2019). Magnetometer-Based Drift Correction During Rest in IMU Arm Motion Tracking. *Sensors*, *19*(6), Article 6. https://doi.org/10.3390/s19061312

Xu, Y., Zhang, J., Zhang, Q., & Tao, D. (2022). ViTPose: Simple Vision Transformer Baselines for Human Pose Estimation. *Advances in Neural Information Processing Systems*, *35*, 38571–38584.

Zhang, S., Li, X., Hu, C., Xu, J., & Liu, H. (2024). DSTFormer: 3D Human Pose Estimation with a Dual-scale Spatial and Temporal Transformer Network. *2024 International Conference on Advanced Robotics and Mechatronics (ICARM)*, 484–489. https://doi.org/10.1109/ICARM62033.2024.10715863

Zhao, P., & Krebs, H. I. (2024). Enabling Home Rehabilitation with Smartphone-Powered Upper Limb Training. *2024 10th IEEE RAS/EMBS International Conference for Biomedical Robotics and Biomechatronics (BioRob)*, 438–443. https://doi.org/10.1109/BioRob60516.2024.10719743

Zheng, C. E., Wu, W., Chen, C., Shah, M., Zheng, C., Yang, T., Zhu, S., Shen, J., & Kehtarnavaz, N. (2018). Deep Learning-Based Human Pose Estimation: A Survey. *J. ACM*, *37*(111), 45469. https://doi.org/10.1145/1122445.1122456

Zheng, C., Zhu, S., Mendieta, M., Yang, T., Chen, C., & Ding, Z. (2021). *3D Human Pose Estimation With Spatial and Temporal Transformers*. 11656–11665. https://openaccess.thecvf.com/content/ICCV2021/html/Zheng_3D_Human_Pose_Estimation_With_Spatial_and_Temporal_Transformers_ICCV_2021_paper.html

Zhu, W., Wu, W., & Wang, Y. (2022). MotionBERT: Unified Pretraining for Human Motion Analysis. *arXiv*.

**Supplementary Material**

All tables, graphs and figures generated are available in the complementary Zenodo repository complementary to this work (Medrano-Paredes et al., 2025). The "analysis" folder contains both the joint angle plots and the angle signals filtered and synchronized with the IMU signals. The "pose3d" folder contains the pose estimation data from video. The "jointangles" folder contains the angles calculated from the pose estimation as well as the angles calculated from the IMUs. The "results" folder contains the comparative plots and tables derived from the evaluation metrics.

*Table 1 Markers used to compute each angle*

| Angle | Marker 1 | Marker 2 | Marker 3 | Marker 4 |
|---|---|---|---|---|
| *arm_flex_r* | right_shoulder | right_elbow | neck | torso |
| *arm_flex_l* | left_shoulder | left_elbow | neck | torso |
| *elbow_flex_r* | right_shoulder | right_elbow | right_elbow | right_wrist |
| *elbow_flex_l* | left_shoulder | left_elbow | left_elbow | left_wrist |
| *knee_angle_r* | right_hip | right_knee | right_knee | right_ankle |
| *knee_angle_l* | left_hip | left_knee | left_knee | left_ankle |

*Table 2 Angle evaluated for each activity*

| Activity IDs | Angle used |
|---|---|
| A01, A03 | *knee_angle_l* |
| A02, A04 | *knee_angle_r* |
| A06 | *elbow_flex_l* |
| A05, A09 | *elbow_flex_r* |
| A08, A12 | *arm_flex_l* |
| A07, A10, A11, A13 | *arm_flex_r* |

*Table 3 Overall evaluation metrics*

| Model | RMSE | MAE | NRMSE | Correlation | $R^2$ |
|---|---|---|---|---|---|
| **BodyTrack** | 10.89 ± 3.67 | 9.00 ± 3.12 | 0.17 ± 0.05 | 0.78 ± 0.12 | 0.44 ± 0.31 |
| **MMPose** | 11.04 ± 4.17 | 9.35 ± 3.61 | 0.17 ± 0.05 | 0.84 ± 0.10 | 0.58 ± 0.26 |
| **MotionAGFormer** | 9.27 ± 4.80 | 7.86 ± 4.18 | 0.14 ± 0.06 | 0.86 ± 0.15 | 0.67 ± 0.28 |
| **MotionBERT** | 12.28 ± 4.59 | 10.15 ± 3.86 | 0.20 ± 0.06 | 0.79 ± 0.11 | 0.16 ± 0.50 |

*Table 4* Activity RMSE across models

| ID | Legend | MotionAGFormer | MotionBERT | MMPose | BodyTrack |
|---|---|---|---|---|---|
| A01 | *walk_forward* | 8.12 ± 3.42 | 7.40 ± 2.56 | 6.54 ± 2.64 | 7.07 ± 2.73 |
| A02 | *walk_backward* | 6.34 ± 2.33 | 6.55 ± 2.18 | 5.74 ± 1.96 | 4.72 ± 1.40 |
| A03 | *walk_along* | 5.66 ± 2.72 | 7.09 ± 2.54 | 6.25 ± 2.61 | 6.02 ± 2.26 |
| A04 | *sit_to_stand* | 12.59 ± 7.71 | 5.71 ± 2.10 | 9.53 ± 2.02 | 6.99 ± 2.00 |
| A05 | *move_right_arm* | 6.14 ± 2.64 | 13.47 ± 3.90 | 10.19 ± 2.70 | 13.02 ± 3.02 |
| A06 | *move_left_arm* | 18.22 ± 8.63 | 33.64 ± 7.45 | 19.29 ± 6.37 | 25.70 ± 8.69 |
| A07 | *drink_right_arm* | 5.58 ± 2.86 | 8.92 ± 3.91 | 8.23 ± 3.83 | 15.04 ± 3.49 |
| A08 | *drink_left_arm* | 8.94 ± 3.21 | 8.39 ± 3.14 | 8.49 ± 2.62 | 8.40 ± 2.97 |
| A09 | *assemble_both_arms* | 6.94 ± 5.38 | 10.75 ± 3.57 | 6.53 ± 5.38 | 10.15 ± 4.01 |
| A10 | *throw_both_arms* | 9.80 ± 9.38 | 19.49 ± 13.45 | 17.46 ± 12.86 | 13.01 ± 8.76 |
| A11 | *reachup_right_arm* | 9.90 ± 4.54 | 15.24 ± 6.05 | 19.01 ± 4.78 | 14.41 ± 3.20 |
| A12 | *reachup_left_arm* | 17.28 ± 7.04 | 6.22 ± 2.43 | 8.75 ± 2.86 | 9.49 ± 2.68 |
| A13 | *tear_both_arms* | 4.99 ± 2.48 | 16.79 ± 6.42 | 17.57 ± 3.53 | 7.61 ± 2.55 |

*Table 5* Activity correlation across models

| ID | Legend | MotionAGFormer | MotionBERT | MMPose | BodyTrack |
|---|---|---|---|---|---|
| A01 | walk_forward | 0.83 ± 0.12 | 0.87 ± 0.10 | 0.88 ± 0.11 | 0.87 ± 0.12 |
| A02 | walk_backward | 0.87 ± 0.11 | 0.80 ± 0.14 | 0.86 ± 0.10 | 0.92 ± 0.04 |
| A03 | walk_along | 0.92 ± 0.06 | 0.89 ± 0.07 | 0.90 ± 0.06 | 0.91 ± 0.05 |
| A04 | sit_to_stand | 0.86 ± 0.33 | 0.98 ± 0.02 | 0.96 ± 0.03 | 0.97 ± 0.02 |
| A05 | move_right_arm | 0.92 ± 0.06 | 0.79 ± 0.11 | 0.87 ± 0.06 | 0.64 ± 0.19 |
| A06 | move_left_arm | 0.20 ± 0.48 | -0.22 ± 0.26 | -0.04 ± 0.36 | -0.17 ± 0.36 |
| A07 | drink_right_arm | 0.98 ± 0.02 | 0.93 ± 0.07 | 0.97 ± 0.03 | 0.74 ± 0.21 |
| A08 | drink_left_arm | 0.91 ± 0.13 | 0.94 ± 0.06 | 0.93 ± 0.05 | 0.97 ± 0.02 |
| A09 | assemble_both_arms | 0.83 ± 0.18 | 0.61 ± 0.18 | 0.87 ± 0.10 | 0.59 ± 0.21 |
| A10 | throw_both_arms | 0.89 ± 0.31 | 0.80 ± 0.28 | 0.88 ± 0.26 | 0.86 ± 0.25 |
| A11 | reachup_right_arm | 0.99 ± 0.01 | 0.97 ± 0.02 | 0.97 ± 0.02 | 0.94 ± 0.02 |
| A12 | reachup_left_arm | 0.96 ± 0.07 | 0.99 ± 0.01 | 0.99 ± 0.01 | 0.99 ± 0.01 |
| A13 | tear_both_arms | 0.98 ± 0.02 | 0.88 ± 0.10 | 0.93 ± 0.05 | 0.94 ± 0.06 |

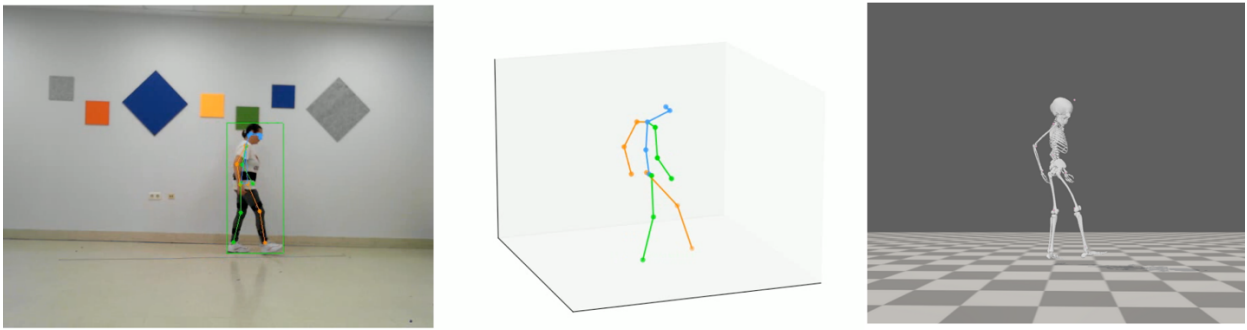

**Fig. 1** Example processing of activity A01 (walk_forward) for subject S40 using 2D-to-3D pose lifting. First, bounding box detection and pose estimation are performed on the original video. A 3D reconstruction of the predicted keypoints can then be visualized. Finally, an inverse kinematics analysis is conducted in OpenSim

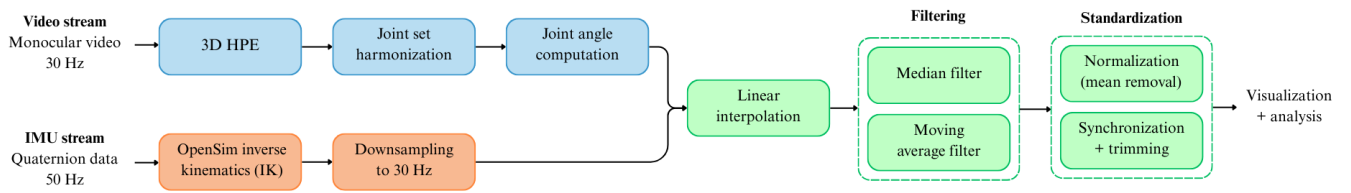

**Fig. 2** End-to-end processing pipeline for video and IMU inverse kinematics

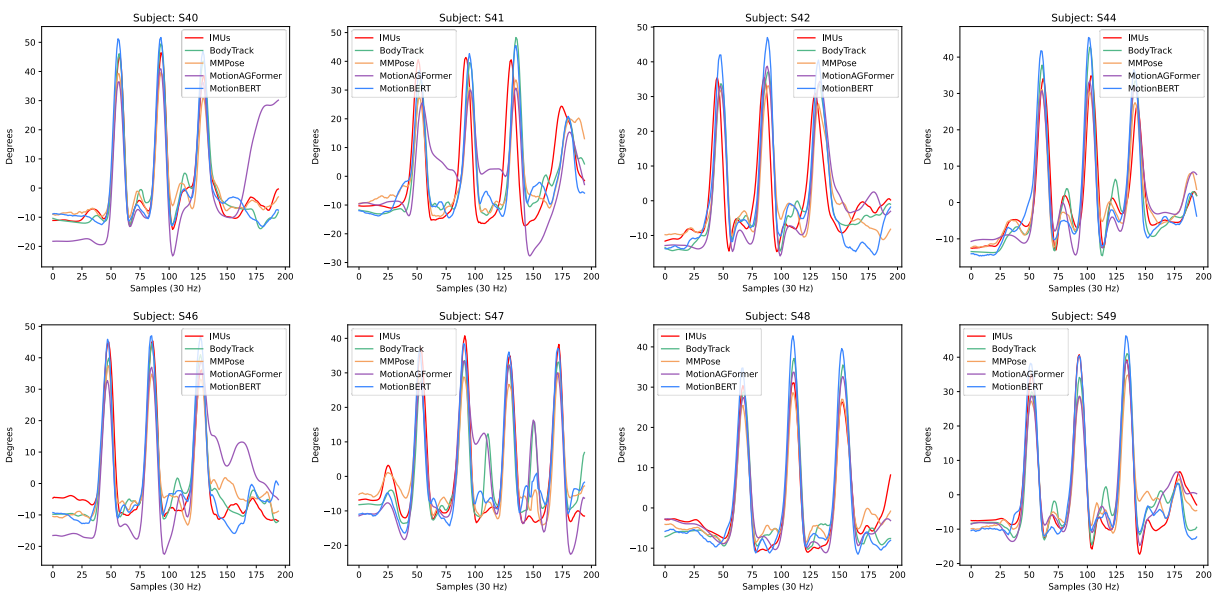

**Fig. 3** *Synchronized and smoothed joint angle knee_angle_l processed by each model for activity A01 (walk_forward). Equivalent plots for every activity and subject can be found in the companion Zenodo repository*

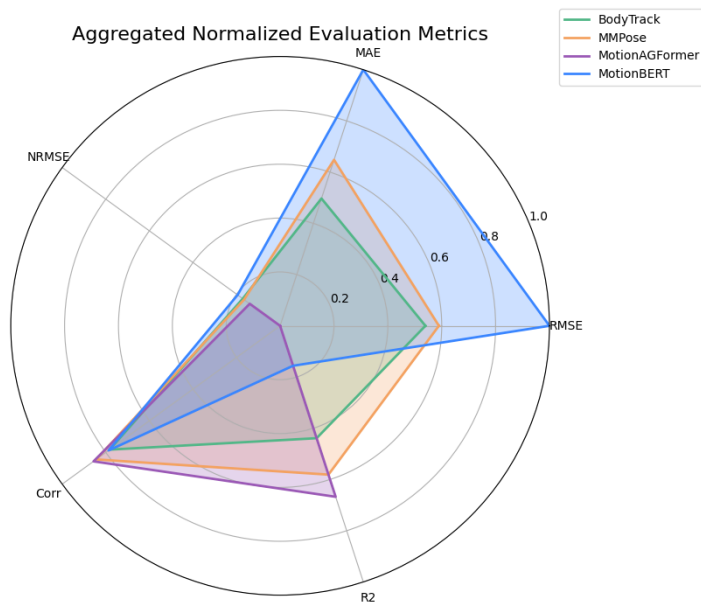

**Fig. 4** *Aggregated and normalized evaluation metrics for all subjects and all activities. The metrics are normalized to the interval [0,1] using min-max scaling. Better performance implies lower RMSE, NRMSE and MAE values, and higher correlation and $R^2$ values*

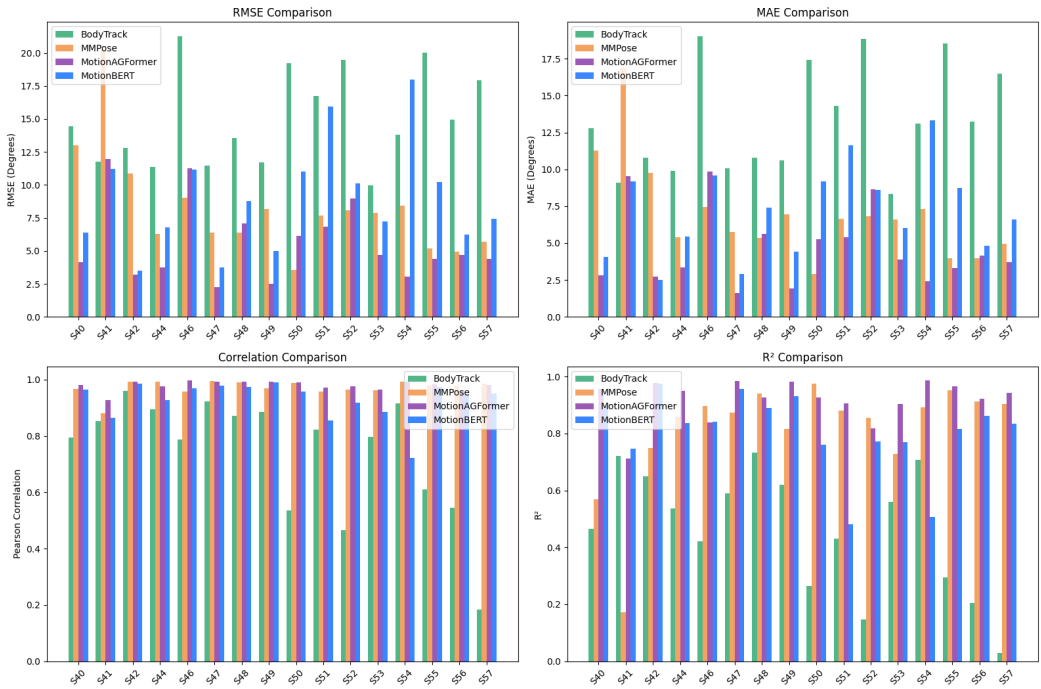

***Fig. 5*** *Evaluation metrics per subject and model for activity A07 (drink_right_arm). Equivalent plots for every activity and subject can be found in the companion Zenodo repository*